# Quantum-Classical Hybrid Framework for Zero-Day Time-Push GNSS Spoofing Detection

Abyad Enan, *Student Member, IEEE*, Mashrur Chowdhury, *Senior Member, IEEE*, Sagar Dasgupta, and Mizanur Rahman, *Senior Member, IEEE*

*Abstract*—Global Navigation Satellite Systems (GNSS) are critical for a wide range of applications that rely on Positioning, Navigation, and Timing (PNT) solutions. However, GNSS are highly vulnerable to spoofing attacks, where adversaries transmit counterfeit signals to deceive receivers into interpreting them as legitimate, thereby falsifying positioning and timing information. Such attacks can lead to severe consequences, including misdirected navigation, compromised data integrity, and operational disruptions. Most existing spoofing detection methods depend on supervised learning techniques that assume access to both authentic and spoofed GNSS signal features. This assumption limits their effectiveness, as they struggle to detect novel, evolved, and unseen attacks. To overcome this limitation, we develop a zero-day spoofing detection method using a Hybrid Quantum-Classical Autoencoder (HQC-AE), trained solely on authentic GNSS signals without exposure to spoofed data. Our approach leverages a set of features that can be extracted during the tracking stage of GNSS signal processing, before the PNT solution is computed, enabling proactive spoofing detection across various receiver platforms. We focus on spoofing detection in static GNSS receivers, which are particularly susceptible to time-push spoofing attacks, where attackers manipulate timing information to induce incorrect time computations at the receiver. We evaluate our model against different unseen time-push spoofing attack scenarios: simplistic, intermediate, and sophisticated. Our analysis demonstrates that the HQC-AE consistently outperforms its classical counterpart, traditional supervised learning-based models, and existing unsupervised learning-based methods in detecting zero-day, unseen GNSS time-push spoofing attacks, achieving an average detection accuracy of 97.71% with an average false negative rate of 0.62% (when an attack occurs but is not detected). For sophisticated spoofing attacks, the HQC-AE attains an accuracy of 98.23% with a false negative rate of 1.85%. These findings highlight the effectiveness of our method in proactively detecting zero-day GNSS time-push spoofing attacks across various stationary GNSS receiver platforms.

## I. INTRODUCTION

### A. Background

GLOBAL Navigation and Satellite Systems (GNSS) are satellite constellations widely used to provide Positioning, Navigation, and Timing (PNT) services across numerous industries and applications [1], [2], [3], [4]. Each system consists of satellites orbiting the Earth at high altitudes, broadcasting signals that ground-based receivers process to determine their location, velocity, and time [5]. PNT solutions derived from GNSS are critical to applications that require precise localization and timing [5], such as transportation [6], telecommunications [7], power grid synchronization [8], financial systems [9], geodesy and surveying [10], scientific research [11], emergency response and rescue services [12], and defense operations [13]. The major operational constellations include the Global Positioning System (GPS) by the United States, the GLONASS by Russia, Galileo by the European Union, and BeiDou by China [14].

Despite their ubiquity, GNSS signals are inherently vulnerable to spoofing attacks due to two factors: their weak signal power when received at Earth's surface [15], [16], [17], and their open, publicly available structure [18] which includes signals and navigation message formats that can be readily replicated by an adversary. Civilian GNSS signals are unencrypted and do not require authentication, allowing adversaries to transmit counterfeit signals with comparable or greater power than the authentic satellite transmissions. Such spoofed signals can override the legitimate ones, causing receivers to lock onto spoofed signals [17], [19], [20], [21]. This leads to erroneous calculations of position or time, posing severe risks to safety and mission-critical applications [22], [23], [24].

Spoofing attacks typically target either the receiver's position (position-push spoofing) or its timing information (time-push spoofing) [13], [19]. Position-push spoofing aims to deceive a GNSS receiver into computing incorrect locations [6], [13], [25]. These attacks have been extensively studied due to their impact on aviation [26], maritime navigation [27], autonomous vehicles [23], and unmanned aerial vehicles

This paragraph of the first footnote will contain the date on which you submitted your paper for review, which is populated by IEEE. This work is based upon the work supported by the National Center for Transportation Cybersecurity and Resiliency (TraCR) (a US Department of Transportation National University Transportation Center) headquartered at Clemson University, Clemson, South Carolina, USA. *(Corresponding author: Abyad Enan).*

Abyad Enan and Mashrur Chowdhury are with the Glenn Department of Civil Engineering, Clemson University, Clemson, SC 29634, USA (e-mail: aenan@clemson.edu; mac@clemson.edu).

Sagar Dasgupta and Mizanur Rahman are with the Department of Civil, Construction, and Environmental Engineering, University of Alabama, Tuscaloosa, AL 35487 USA (e-mail: sdasgupta@ua.edu; mizan.rahman@ua.edu).

(UAVs) [23]. Time-push spoofing, on the other hand, alters GNSS-derived time [28] and disrupts systems that depend on precise synchronization, leading to widespread disruptions [28]. Time-push spoofing is especially effective against static receivers, which lack motion-induced validation mechanisms and rely solely on GNSS for timing, not positioning. Such receivers are commonly embedded in fixed infrastructure, including base stations, data centers, and grid substations. By gradually shifting perceived time, attackers can introduce synchronization errors without triggering alarms, potentially propagating failures across interconnected systems [19]. While position-push spoofing has received significant attention, time-push spoofing is equally critical, particularly in the context of static receivers that remain stationary and rely primarily on GNSS for accurate timing.

*B. Motivations*

Figure 1 illustrates how various critical systems rely on GNSS for precise timing, with the time distributed via the Network Time Protocol (NTP) over the Internet. The entire ecosystem illustrates the widespread dependence on GNSS-derived timing, highlighting the potential cascading impact of a timing disruption or spoofing attack across interconnected components.

Distributed systems that rely on coordinated sensing and control are especially vulnerable to such time-push spoofing attacks. For instance, in the Internet of Things (IoT) ecosystem, many sensors and actuators depend on GNSS-derived time for synchronized data collection, logging, and actuation [29], [30]. A time-push spoofing attack can misalign timestamps across these devices, undermining the consistency of event logs and compromising time-sensitive actions. In smart agriculture, this might cause irrigation systems to activate at incorrect times, leading to crop damage. In smart factories, temporally misaligned logs could confuse predictive maintenance algorithms, increasing the risk of undetected mechanical issues and system downtime [29].

Autonomous vehicles (AVs), particularly those on the Internet-of-Vehicles (IoV) ecosystem, are an integral part of the broader IoT. When an AV is stationary, such as when stopped at intersections or waiting in parking lots, it becomes vulnerable to time-push spoofing attacks. AVs rely on GNSS-derived time to synchronize system clocks and align sensor data streams, enabling seamless operation and coordination with external infrastructure [31], [32]. Successful time-push spoofing during idle periods can distort system time, disrupt data logging, delay critical functions, or desynchronize a vehicle from surrounding infrastructure, increasing the risk of navigation errors and safety issues once the vehicle is in operation [33]. Furthermore, AVs interact with roadside units (RSUs) that support vehicle-to-infrastructure (V2I) communication and traffic coordination mechanisms [34], such as green-wave signaling and emergency vehicle prioritization. RSUs also depend on precise timing to maintain message integrity. If either an AV or an RSU is not time-synchronized due to a time-push spoofing attack, it can result in authentication failures and disrupt cooperative traffic management approaches. Since these systems are closely interconnected, a timing disruption in one component can propagate throughout the broader vehicular network ecosystem.

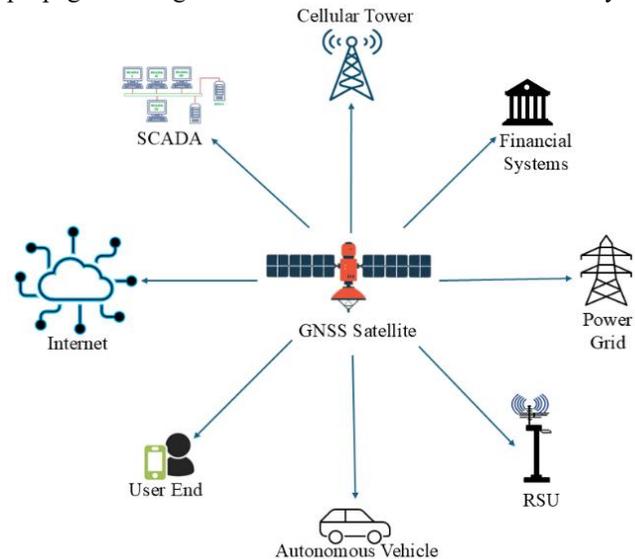

**Fig. 1.** Interconnected cyber-physical infrastructure reliant on GNSS-based time synchronization and application-level connectivity among components.

The impact of time-push spoofing extends beyond transportation and into broader communication systems. For example, 5G networks and other time-division multiplexed communication channels rely on GNSS-based timing to maintain synchronization between base stations and user devices [35]. A time-push spoofing attack that manipulates GNSS signals at a communication base station can degrade signal quality, cause dropped calls, or delay handover disruptions that may also affect any system dependent on stable, low-latency communication, including connected vehicles and time-critical industrial processes. These communication base stations are an integral part of the IoT infrastructure, as they are essential for enabling wireless communication between IoT devices and the broader network. Specifically, communication base stations, often deployed as cell towers, serve as the physical structures that transmit and receive signals, facilitating the connection of IoT devices to cellular networks and the internet. A time-push spoofing attack that compromises time synchronization at the communication base station can propagate through the entire IoT ecosystem, disrupting dependent applications and services.

Beyond transportation and communication, maintaining timing integrity is equally vital for many other critical infrastructures. Supervisory Control and Data Acquisition (SCADA) systems used in water treatment plants, energy distribution hubs, and transportation nodes rely on precise timing for scheduled operations, logging, and control execution [36]. A gradual time drift caused by time-push spoofing can corrupt operational records, trigger false alarms, or cause execution delays, complicating incident diagnosis and degrading system resilience over time [36], [37]. Power systems, in particular, represent a domain where even minor time synchronization errors can have severe consequences. Phasor measurement units (PMUs) and substations rely on tightly synchronized GNSS time for wide-area monitoring, fault detection, and state estimation [8], [38]. A time-push attack can introduce timestamp errors in PMU data, leading to corrupted synchro phasor measurements and false assessments of grid stability. This desynchronization, especially under stressed conditions, can prompt incorrect control actions, mismanaged load distribution, or even cascading blackouts [38]. As the power grid evolves toward smart grid architectures, with increasing automation and data-driven control, the sensitivity to precise time synchronization becomes even more pronounced.

Financial systems are another critical sector where GNSS-aided time synchronization plays a foundational role. Global trading platforms, stock exchanges, and transaction auditing systems rely on accurate time to maintain the order of trades and ensure compliance with financial regulations [39]. In high-frequency trading environments, where microsecond-level precision is essential, a time-push attack could reorder trades, create discrepancies during reconciliation, or result in unfair advantages and regulatory violations [40]. Their reliance on distributed infrastructure and time-synchronized processes positions them as vulnerable to the broader implications of spoofing-based timing attacks [39].

Taken together, these examples highlight how time-push spoofing exploits a critical dependency in modern digital systems: the trust placed in GNSS time synchronization. Whether affecting distributed sensors, vehicles, communication backbones, industrial control systems, power networks, or financial markets, the attack undermines the temporal foundation upon which these systems operate. Some of these domains are explicitly interconnected, share the same vulnerability at the GNSS receiver level. Time synchronization, if often taken for granted, may become a single point of failure, and its compromise can trigger faults that are difficult to trace but devastating in consequence.

*C. Focus of This Study*

To protect critical infrastructures, it is essential to detect GNSS spoofing attacks proactively, before they lead to incorrect positioning or timing solutions. Numerous Machine Learning (ML) and Deep Learning (DL)-based methods have been developed to detect GNSS spoofing attacks; however, these approaches often rely on the assumption that the detection system has prior knowledge of the spoofing characteristics. As GNSS spoofing techniques continue to evolve, models trained on specific spoofing patterns may fail to generalize to previously unseen attack types. Consequently, only limited efforts have been made to develop spoofing detection methods that learn solely from authentic signals, enabling the detection of novel spoofing attacks without prior exposure.

In this study, we address these limitations by developing an unsupervised learning-based zero-day (i.e., previously unseen by a spoofing detector) time-push spoofing attack detection method tailored for GNSS static receivers. Our approach employs a hybrid quantum-classical autoencoder (HQC-AE), leveraging the computational advantages of quantum computing to proactively detect zero-day time-push spoofing attacks, without having any prior knowledge about spoofed signals. Experimental results on benchmark spoofing scenarios demonstrate that our HQC-AE outperforms classical autoencoders (AEs), traditional supervised learning methods, and existing unsupervised learning approaches from the literature in identifying previously unseen attacks.

The remainder of this manuscript is organized as follows: Section II reviews related studies from the literature. Section III outlines the contributions of this study and explains how our method addresses the limitations of existing approaches. Section IV presents the attack model, while Section V describes the dataset preparation process based on the defined attack model. Section VI details our detection method. Section VII presents the analysis and results. Finally, Section VIII draws the conclusions.

## II. RELATED STUDIES

A wide range of GNSS spoofing detection methods has been reported in the literature, including single-antenna, multi-antenna, inertial, and other sensor-based approaches [16]. Among these, single-antenna methods are often preferred for their cost-effectiveness, practicality, and ease of deployment, as they require no additional hardware. The single antenna-based methods monitor signal power directly [41] or using an automatic gain control (AGC) [42], an autocorrelation profile of the receiver [43], [44], [45], [46], and a combination of both [16], [47]. These methods utilize a Bayesian detection framework for spoofing attack detection. However, Bayesian methods are not always reliable for unseen or novel spoofing attacks due to their dependence on accurate prior knowledge and statistical models. Specifically, the detection performance relies heavily on well-characterized likelihood functions for both genuine and spoofed signals. If the feature distributions under spoofed conditions are not well modeled, particularly for spoofing strategies that differ from those seen during training, the resulting likelihood ratios or posterior probabilities can be misleading. As a result, the detector may either fail to flag sophisticated or previously unobserved spoofing attacks or raise false alarms in benign conditions that do not match the assumed models.

Different ML and DL-based methods are developed for spoofing attack detection. The advantage of using ML and DL methods is that they do not require explicit mathematical modeling of signal behavior or noise characteristics, allowing them to learn complex, nonlinear patterns directly from data. In [48], the authors developed an Artificial Neural Network (ANN) to detect spoofing attacks using pseudorange, carrier phase, carrier-to-noise (C/N₀) and Doppler shift. In [49], the authors utilized 13 tracking features along with position, velocity and time (PVT) blocks for multiple ML-based classifier-based detection methods. In [50], the authors used signal power and delta and early-late phase signal quality monitoring (SQM) metrics to implement detection methods utilizing ANN, Naive

Bayes, and K-Nearest Neighbors (K-NN). In [51], the authors developed a spoofing detection method that utilizes the Cross Ambiguity Function (CAF) obtained from the receiver's acquisition stage. They employed an ANN and a Convolutional Neural Network (CNN) to identify spoofing by detecting the presence of multiple peaks within the CAF. All of these approaches rely on supervised learning techniques that require models trained with features from both authentic and spoofed signals. A major challenge with supervised learning is that models often perform poorly when evaluated on data that differs significantly from the training set or contains previously unseen patterns. Additionally, a key limitation of the mentioned ML-based methods is their limited evaluation across diverse spoofing scenarios. Most studies use custom, in-house datasets rather than standardized spoofing datasets or publicly available benchmarks. Therefore, applying ML methods to such standardized datasets would be a valuable direction for further investigation.

A limited number of studies have explored unsupervised learning-based methods for spoofing attack detection, particularly for identifying unseen attacks. The authors in [52] developed a Generative Adversarial Network (GAN) trained on authentic, unspoofed signals to detect such attacks. However, this approach is computationally expensive due to the requirement of computing multiple CAFs. Another limitation is that the method was evaluated using synthetic data rather than realistic spoofing scenarios or standard benchmarks, which restricts its practical applicability.

To overcome the limitations of the GAN-based method in [52], the authors in [53] developed a representation learning-based spoofing attack detection method using a Variational Autoencoder (VAE), where the VAE is trained with clean, unspoofed data. The authors evaluated their method using a spoofing benchmark and achieved up to 92.5% detection accuracy for a sophisticated spoofing scenario with a false negative rate (spoofing attack happened but not detected) of 7.5%. In [54], the authors developed an improved version of the VAE-based method of [53], developed a cascade neural network incorporating VAE along with a GAN to improve the detection accuracy for sophisticated spoofing scenarios and achieved an accuracy of up to 95% with a false negative rate of 5.31% for a sophisticated attack scenario. However, a key limitation of these approaches is their difficulty in balancing the trade-off between false positive and false negative rates, particularly in the case of sophisticated spoofing attacks. Specifically, when the models aim to have a low false positive rate, the false negative rate increases significantly to 13.25% [53] and 23.26% [54], resulting in more spoofed samples being left undetected. This trade-off poses a concern for safety-critical or time-sensitive applications, where the consequences of undetected spoofing can be far more severe than occasional false alarms. Therefore, although the model demonstrates robustness against previously unseen attacks, its detection sensitivity needs to be carefully calibrated to avoid compromising performance under stringent false alarm constraints.

## III. CONTRIBUTIONS

In this study, we address the limitations of existing GNSS spoofing detection methods by developing an HQC-AE to detect time-push spoofing attacks targeting GNSS static receivers. The HQC-AE is trained exclusively on authentic GNSS satellite signals and can detect spoofed signals without any prior knowledge of spoofing characteristics, enabling zero-day detection capability. Our approach prioritizes high detection accuracy while minimizing the false negative rate (FNR), where false negatives occur when spoofed samples are left undetected and incorrectly classified as legitimate. Reducing false negatives is particularly critical, as undetected spoofing attacks can have more severe consequences than false positives, where legitimate activity is mistakenly flagged as spoofed.

To the best of our knowledge, this is the first study to apply quantum neural networks to the detection of GNSS spoofing attacks. We evaluate our method using a realistic GNSS spoofing benchmark that encompasses simplistic, intermediate, and sophisticated attack scenarios. The key contributions of this study are as follows:

- A new approach to combine a set of features that can be extracted during the tracking stage of GNSS signal processing, before the PNT solution is computed, making them suitable for proactive spoofing detection across various receiver platforms.
- Introduce an innovative strategy to integrate quantum artificial intelligence into GNSS with the development of an unsupervised-learning-based time-push spoofing attack detection for GNSS static receivers. Our detection method employs an HQC-AE, capable of detecting unseen spoofing attacks without having any prior knowledge about any GNSS spoofed signals, enabling zero-day detection capability.
- Evaluate our detection method on a GNSS spoofing benchmark comprising diverse, realistic scenarios, and compare its performance against a classical AE, traditional supervised learning models using a leave-one-out approach for unseen attacks, and existing unsupervised-learning methods, demonstrating the superiority, in terms of both accuracy and FNR, of the HQC-AE over all baselines.

## IV. ATTACK MODEL

In the field of GNSS, selecting a standard or benchmark is essential for evaluating spoofing attack detection methods, particularly given the diversity and evolving nature of spoofing scenarios. Since such attacks involve subtle manipulations of signal timing, power, and structure, a robust GNSS signal authentication technique must be tested against standardized spoofing attacks that feature high-fidelity GNSS recordings, wherein spoofed signals interact with authentic signals at the signal level using realistic attack strategies.

To evaluate the performance of our GNSS spoofing attack detection method, we utilize the Texas Spoofing Test Battery (TEXBAT) [19] attack model. Developed by the Radionavigation Laboratory at the University of Texas at Austin, TEXBAT is a set of high-fidelity digital recordings of live GPS spoofing attacks. It is widely recognized as an important resource and benchmark for testing anti-spoofing technologies in civil GPS receivers. TEXBAT is regarded as a cornerstone in the GNSS security research community due to its authenticity, repeatability, and signal-level fidelity. It includes scenarios in which a software-defined GPS spoofer interacts with either real-time or replays authentic GPS signals across various configurations. Because the spoofing attacks in TEXBAT are designed to emulate the capabilities of sophisticated yet plausible adversaries, any receiver that can successfully detect and respond to these scenarios demonstrates a meaningful level of spoof resistance under real-world threat models. Thus, TEXBAT not only provides a basis for comparative performance evaluation but also serves as an evolving benchmark for certifying GPS receivers as spoof-resistant, as acknowledged in congressional testimony [55].

TEXBAT includes six primary spoofing scenarios (ds1 through ds6) [19], with additional scenarios added later (ds7 and ds8) [56]. These scenarios vary in spoofing type (switch-over, time-push, or position-push), power advantage, receiver mobility (static vs. dynamic), and spoofer configuration (e.g., frequency lock mode, carrier phase control). Scenarios ds5 and ds6 involve dynamic receivers, while the remaining scenarios target static receivers. Among the scenarios for static receivers, ds1 represents a switch-over attack; ds2, ds3, and ds7 correspond to time-push attacks; and ds8 implements a security code estimation and replay (SCER) attack. Since our study focuses on the detection of time-push spoofing attacks, we consider scenarios ds2, ds3, and ds7 in this study.

Scenarios ds2, ds3, and ds7 represent progressively sophisticated time-push attacks on static receivers [16], [19]. Scenario ds2 is a brute-force attack with a 10 dB power advantage, code-phase alignment, and no frequency lock, causing observable beating patterns and a constant code-phase pull-off of 1.2 m/s after takeover. Scenario ds3 increases stealth by using matched power (within 1.3 dB), frequency lock, and smoother transitions, minimizing anomalies while maintaining the same code-phase pull-off [16], [19]. Scenario ds7 is the most sophisticated, combining matched power, frequency and carrier-phase alignment, and controlled amplitude modulation to cancel authentic signals without Automatic Gain Control (AGC) or carrier-to-noise ($C/N_0$) anomalies, before initiating a stealthy time-push pull-off [16], [56].

As previously discussed, a wide range of critical applications relies on precise timing information derived from GNSS. TEXBAT time-push spoofing attack scenarios present the ways a receiver's internal clock can be tricked by manipulating the code phase of incoming GPS signals. This deception can lead to gradual but cumulative errors in system time, which in many domains can have catastrophic operational consequences [5], [11], [29], [30], [31], [40], [41], [57]. Given the severity of such consequences, it is essential that time-push spoofing detection mechanisms be evaluated against realistic, signal-level time-push attacks [19], [53], [54], [55]. TEXBAT ds2, ds3, and ds7 attack scenarios demonstrate a plausible and technically feasible attack strategy that could be executed in practice using commercially available software-defined radios and open-source GPS spoofing frameworks. Each scenario also represents distinct levels of stealth and sophistication. The progressive nature of these scenarios ensures that detection methods are evaluated not only for their ability to flag overt spoofing but also for their robustness against subtle, low-observability threats. Importantly, if a spoofing detection method, trained or configured without prior knowledge of these scenarios, can effectively detect spoofed signals in ds2, ds3, and ds7, it is justifiable to conclude that the method has learned generalizable features and should be capable of detecting unknown or novel time-push attacks. In this context, evaluation on TEXBAT ds2, ds3, and ds7 offers a rigorous and representative benchmark for time-push attacks. Successful detection across these cases provides strong justification for the method's applicability in the field and supports its use in protecting critical time-dependent infrastructures from GPS spoofing attacks.

## V. DATASET PREPARATION

In this section, we discuss how the dataset is prepared for this study by processing different scenarios of TEXBAT and the feature selection techniques. The process is illustrated in Fig. 2.

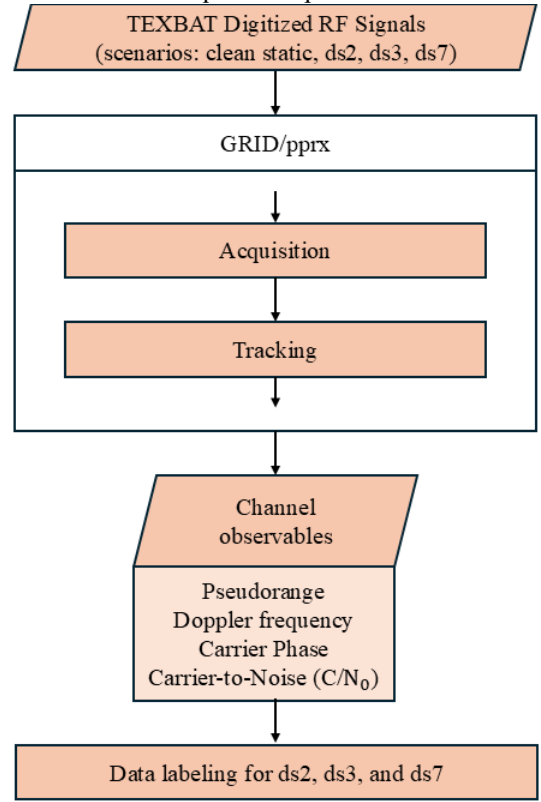

**Fig. 2.** Data preprocessing workflow.

### A. Data Processing Utilized in TEXBAT

We collect the processed TEXBAT data for the unspoofed clean static scenario, and the time-push spoofing attack scenarios, ds2, ds3, and ds7, for static receiver configuration from [58] provided by the Radionavigation Laboratory at the University of Texas at Austin. The following is the summary of the process TEXBAT followed.

The recorded digitized RF signals for clean and spoofed scenarios are processed by the Precise Positioning Receiver (pprx) software receiver [59], developed as part of the University of Texas at Austin's GNSS Receiver Implementation and Development (GRID) framework. GRID/pprx is a post-processing tool designed to extract navigation observables from raw GNSS intermediate-frequency (IF) or baseband signal recordings. It extracts navigation observables and receives state information by emulating the signal processing steps of a receiver.

For each satellite, GRID/pprx performs a two-dimensional search in code delay and Doppler frequency to detect the presence of a signal and estimate its coarse time and frequency offset. This identifies visible satellites and initializes their tracking loops. After acquisition, the software establishes Delay-Locked Loops (DLL) and Phase-Locked Loops (PLL) for code and carrier tracking, respectively. These loops continuously refine the estimates of code phase and carrier frequency. Tracking is performed independently per satellite and produces time-aligned signal replicas for correlation. From the tracked signals, the receiver computes pseudorange, Doppler frequency, carrier phase, and $C/N_0$. These observables are stored in channel.mat and related files, organized per channel (i.e., satellite). GRID/pprx combines pseudorange measurements from multiple satellites to compute the receiver's position and clock offset and is stored in navsol.mat file. GRID/pprx also outputs intermediate data such as in-phase and quadrature (I/Q) samples of the signal after downconversion (iq.mat), filtered and resampled correlator taps (iqtaps.mat), power measurements across selected bandwidths, diagnostic symmetry or correlation metrics (symmdiff.mat), transmitter metadata including spoofing parameters if available (txinfo.mat).

*B. Data Visualization for Feature Selection*

For our detection method, we choose features extracted from the channel.mat file. These features include pseudorange, Doppler frequency, carrier phase, and $C/N_0$ across different channels, which are obtained during the tracking stage of GNSS signal processing, prior to the computation of the PNT solution. Because these features are available early in the signal processing chain, they enable the proactive detection of spoofing attacks before compromised PNT outputs are produced. In contrast to raw intermediate-frequency (I/Q) data or high-rate correlator taps, which require substantial computational and storage resources [60], [61], channel-level observables are lightweight and suitable for real-time analysis [62]. While some prior studies have utilized raw I/Q samples for spoofing detection due to their fine-grained signal fidelity, such approaches typically demand high processing throughput [60], [61]. Besides, channel observables can be derived from any receiver [62]; therefore, using these features offers a practical trade-off between early detection capability and computational feasibility, particularly for receiver-agnostic or resource-constrained implementations.

We analyze the features: $C/N_0$, Doppler frequency, and carrier phase for all observable channels to see if they deviate for the spoofing scenarios from the clean samples. First, we plot the $C/N_0$ observations for clean and spoofing scenarios for all the channels. We observe that at first, the pattern looks the same for all the scenarios across all the channels. This is expected, as no spoofing occurs during the first ~100 seconds in any spoofing scenario [19]. However, once spoofed signals are introduced, the $C/N_0$ values begin to deviate from those observed in the clean static scenario. To illustrate this deviation, we present the plot for the channel corresponding to pseudorandom noise (PRN) 3 in Fig. 3. A PRN refers to a unique code sequence assigned to each GPS satellite, allowing receivers to distinguish between signals from different satellites. This example highlights the deviation of $C/N_0$ from the clean authentic signal under spoofing conditions. The deviation is most pronounced in scenario ds2, which benefits from a 10 dB power advantage. In ds3, a significant deviation is observed initially, but the spoofed signal subsequently attempts to align with the clean static pattern. In contrast, ds7 represents the most sophisticated and stealthy spoofing strategy, closely matching the clean scenario with minimal deviations compared to ds2 and ds3 [56].

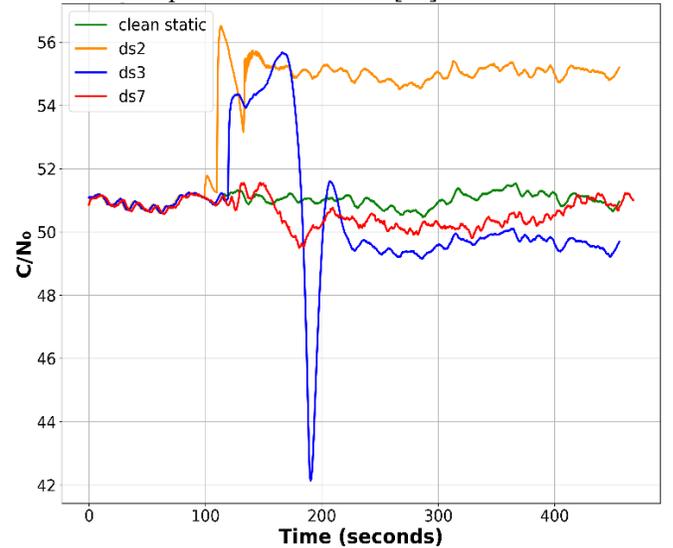

**Fig. 3.** Comparison of $C/N_0$ for clean and different time-push spoofing attack scenarios for the static receiver configuration for the channel with PRN 3.

A similar trend is observed for the Doppler frequency. The deviation is most pronounced across all channels in scenario ds2, as the spoofed signal is not frequency locked. In contrast, since the spoofed signals in ds3 and ds7 are frequency-locked, the Doppler frequency does not exhibit significant deviation in those cases. To illustrate this, we present a comparison of Doppler frequency values over time for PRN 20 in Fig. 4.

We also observe the beat carrier phase for all the channels, which shows slight deviations on some of the channels. Fig. 5

illustrates the beat carrier phase for the channel with PRN 138. Additionally, we investigate pseudorange across all the channels and find slight differences between different spoofing scenarios.

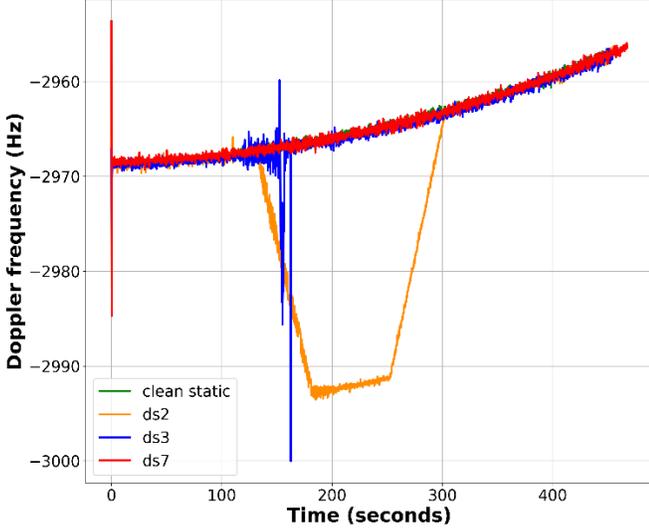

**Fig. 4.** Comparison of Doppler frequency for clean and different time-push spoofing attack scenarios for the static receiver configuration for the channel with PRN 20, where clean static is overlapped with other scenarios.

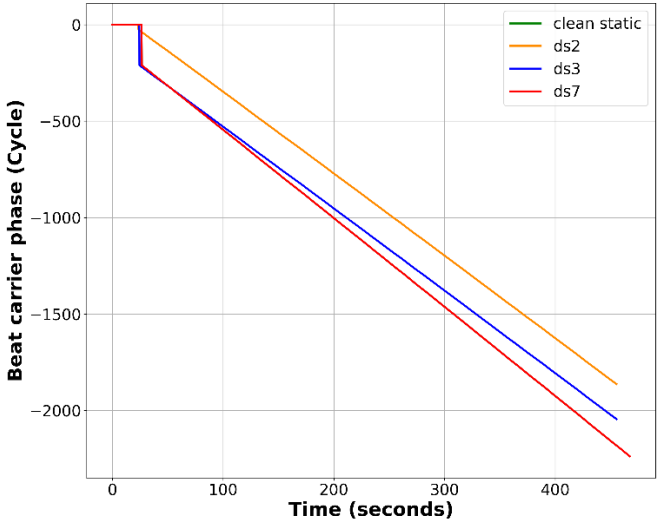

**Fig. 5.** Comparison of beat carrier phase for clean and different time-push spoofing attack scenarios for the static receiver configuration for the channel with PRN 138.

We select the above-mentioned four features, as they exhibit distinguishable patterns under clean and spoofed scenarios, from 13 channels, identified by their respective PRNs, based on the availability of channel observables across all scenarios considered. This results in a total of 52 features. Our model is trained with 52 features for clean static data, which is explained later in Section VI. Each spoofing scenario, ds2, ds3, and ds7, contains both clean and spoofed signals, as no spoofing is introduced during the initial ~100 seconds of the recordings [19]. The presence of both authentic and spoofed signals in these test scenarios enables us to evaluate whether the detection method, trained exclusively on clean signals, can successfully distinguish between genuine and spoofed observations.

## VI. ATTACK DETECTION METHOD

To develop an unsupervised learning framework for a zero-day time-push spoofing attack detection, we design an HQC-AE that is trained using features extracted from clean GNSS signals. By leveraging the representational power of hybrid quantum-classical neural networks, the HQC-AE is expected to effectively learn the underlying features of clean signals, potentially offering a quantum advantage when learning those features. The HQC-AE is tasked with learning a lower-dimensional representation of the input data by encoding it into a compressed format and then decoding it to reconstruct the original input. The difference between the input and its reconstruction, referred to as the reconstruction loss or error, serves as a key indicator in our detection strategy.

When trained solely on authentic (i.e., clean static) data, the HQC-AE becomes specialized in reconstructing such inputs. As a result, it struggles to accurately reconstruct data that deviates from the features of clean samples, such as spoofed signals, leading to a higher reconstruction loss. A threshold for this loss is determined during the training phase. If the reconstruction loss for a given input exceeds this predefined threshold, the input is classified as anomalous, which in our context corresponds to a spoofed sample. This forms the basis of our unsupervised spoofing detection approach, trained only on clean data. The spoofing detection method is illustrated in Fig. 6. In the following subsections, we detail the detection methodology by describing the architectures of both the classical AE, which serves as a classical benchmark, and our HQC-AE, along with the associated detection process.

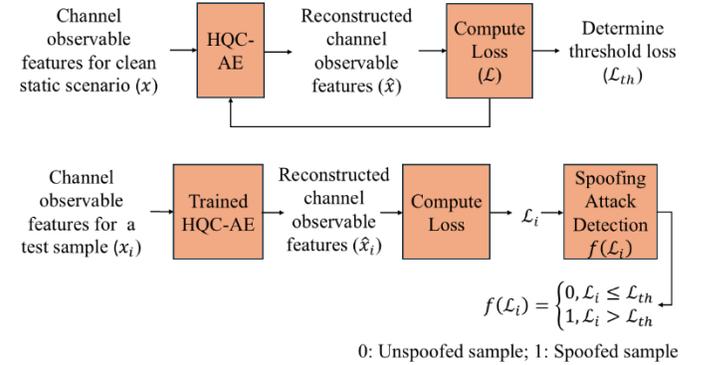

**Fig. 6.** Overview of HQC-AE-based GNSS time-push spoofing attack detection method.

### A. Classical Autoencoder (AE) Architecture

We implement a symmetric feedforward AE composed of fully connected layers for feature learning and dimensionality reduction. The architecture consists of an encoder and a decoder, each with three layers, designed to reduce an input vector of dimensionality = 52, equivalent to the number of input data or features, to a compact latent representation of size 16, and then reconstruct it back to the original input dimension, as shown in Fig. 7. We experiment with AEs of varying architectures by adjusting the number of layers, nodes, and latent space size, and present the best-performing architecture in Fig. 7.

The encoder network performs a gradual compression of the input vector through successive nonlinear transformations. It comprises three fully connected layers with ReLU activations applied to the hidden layers. In layer 1, a linear transformation projects the input vector from $R^{52}$ to $R^{64}$ followed by a ReLU activation. Similarly, a linear transformation from $R^{64}$ to $R^{32}$ is performed in layer 2, followed by a ReLU activation. The final linear layer of the encoder projects to the latent space of dimension $R^{16}$, without any activation function. Formally, the encoder function $f: R^{52} \rightarrow R^{16}$ can be expressed as:

$$z = f(x) = W_3. ReLU(W_2. ReLU(W_1. x + b_1) + b_2) + b_3 \quad (2)$$

where, $x \in R^{52}$ is the input vector, $z \in R^{16}$ is the latent representation, and $W_i b_i$ are the weight matrices and biases for the $i^{th}$ layer.

The decoder mirrors the encoder structure to reconstruct the input from its latent representation. It also consists of three fully connected layers. Layer 1 performs a linear transformation from $R^{16}$ to $R^{32}$, followed by a ReLU activation. Similarly, layer 2 performs a linear transformation from $R^{32}$ to $R^{64}$, followed by a ReLU activation. The final layer projects back to $R^{52}$, with no activation function to allow unrestricted reconstruction. The decoder function $g: R^{16} \rightarrow R^{52}$ can be expressed as follows, where $\hat{x}$ is the reconstructed output:

$$\hat{x} = g(z) = W_6. ReLU(W_5. ReLU(W_4. z + b_4) + b_5) + b_6 \quad (3)$$

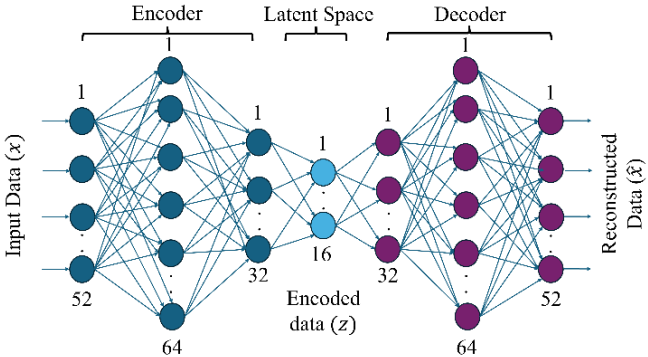

**Fig. 7.** Architecture of classical autoencoder (AE).

*B. Hybrid Quantum-Classical Autoencoder (HQC-AE) Architecture*

To leverage the expressive power of quantum computing feature space, we develop a hybrid autoencoder architecture that integrates quantum computing principles within a classical neural network framework. The architecture consists of a classical encoder, a quantum latent transformation layer, and a classical decoder, as shown in Fig. 8(a). This design enables the nonlinear transformation of latent representations using quantum circuits, enhancing the representation capacity of the latent space.

The classical encoder consists of two fully connected (FC) layers that reduce the dimensionality of the input vector to a latent representation of size 16. In layer 1, a linear transformation projects the input vector from $R^{52}$ to $R^{64}$ followed by a ReLU activation, and layer 2 then projects to the classical latent space of dimension $R^{16}$, without any activation function. The encoder function can be expressed as:

$$z = f(x) = W_2. ReLU(W_1. x + b_1) + b_2 \quad (4)$$

where, $z \in R^{16}$ represents the classical latent space.

To enhance the representational capacity of the latent space, we embed a variational quantum circuit (VQC) between the classical encoder and decoder. VQC is a widely used technique in quantum-classical hybrid learning frameworks where a parametrized quantum circuit processes data or encodes problem instances, and a classical optimizer iteratively updates its parameters to minimize a task-specific cost function [63]. This quantum layer serves as a nonlinear transformation applied to the encoded latent representation, enabling the model to project the input into a quantum-enhanced feature space. The quantum transformation consists of three main stages: amplitude embedding, entangling layers, and measurement, as shown in Fig. 8(b).

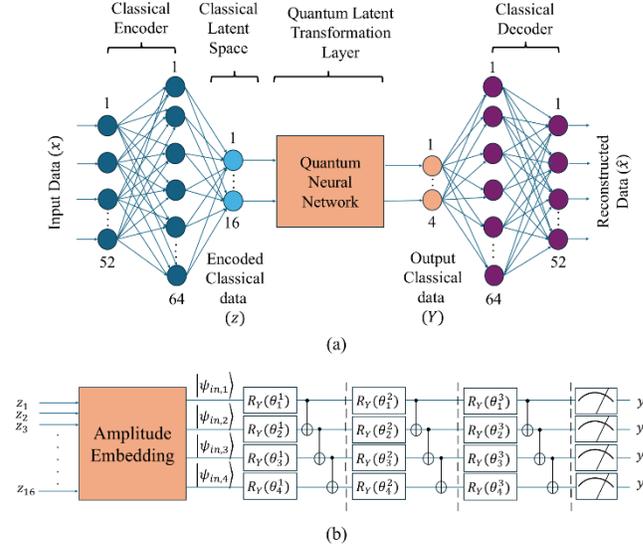

**Fig. 8.** Architecture of (a) Hybrid Quantum-Classical Autoencoder (HQC-AE), (b) quantum neural network.

The classical 16-dimensional latent vector z is mapped to a quantum feature space using amplitude embedding with four qubits. Since amplitude embedding requires $n$ qubits for $2^n$ features, in our case, for 16 or $2^4$ features, we require four qubits. The latent vector from the classical encoder is normalized and encoded into a quantum state using the following equation:

$$|\psi_{in}(z)\rangle = \sum_{i=0}^{15} z_i |i\rangle \quad (5)$$

Where the vector is first normalized to satisfy the quantum state's unit norm constraint using the following equation:

$$\sum_{i=0}^{15} z_i^2 = 1 \quad (6)$$

The amplitude embedding in (5) maps classical information into the quantum Hilbert space, where quantum computation can perform transformations that are not linearly accessible in classical spaces. To apply a learnable transformation in the quantum space, we employ a parameterized quantum circuit composed of three layers of the BasicEntanglerLayers template by PennyLane [64]. This template applies to a stack of rotation and entangling gates across qubits to form a layered VQC. Each layer consists of single-qubit rotations $R_Y(\theta_i)$ on each qubit. Entangling operations are applied in a ring topology between neighboring qubits using CNOT (Controlled-NOT) gates, a fundamental two-qubit quantum logic gate used in quantum computing. The overall unitary transformation applied by the variational circuit is denoted as:

$$U(\theta) = \prod_{l=1}^{3} U_l(\theta_l) \qquad (7)$$

where $\theta$ are trainable parameters, optimized during training, and $U_l$ represents the unitary operator for the $l^{\text{th}}$ layer. The state after quantum operations is:

$$|\psi_{out}\rangle = U(\theta)|\psi_{in}(z)\rangle \qquad (8)$$

To extract classical information from the final quantum state, we perform expectation value measurements of the Pauli-Z operator on each qubit, which can be expressed as:

$$y_i = \langle \psi_{out}|Z_i|\psi_{out}\rangle; \text{ for } i = 0, \dots, 3 \qquad (9)$$

The quantum state is measured in the Z-basis, resulting in a 4-dimensional output vector, $Y = [y_0, y_1, y_2, y_3]^T \in R^4$.

The decoder reconstructs the original input from the quantum-processed latent representation using two fully connected classical layers. Layer 1 performs a linear transformation from $R^4$ to $R^{64}$, followed by a ReLU activation. The final layer projects back to $R^{52}$, with no activation function. The decoder can be expressed as:

$$\hat{x} = h(Y) = W_4 . ReLU(W_3 . Y + b_3) + b_4 \qquad (10)$$

*C. Model Training and Attack Detection*

Both classical AE and HQC-AE models are trained on the clean static data. The input features are standardized using z-score normalization. A validation set, 20% of the training data, which means clean static data, is used to monitor the reconstruction loss of the trained models on the validation set in each epoch. The models are trained minimizing the reconstruction error between the original input $x$ and the reconstructed output $\hat{x}$ using the Mean Squared Error (MSE) loss function as follows:

$$\mathcal{L}(x, \hat{x}) = \|x - \hat{x}\|_2^2 \qquad (11)$$

The training is performed using the Adam optimizer for both classical AE and HQC-AE models. For the HQC-AE, the quantum circuit parameters are treated as trainable PyTorch parameters via PennyLane's TorchLayer interface [64]. During backpropagation, quantum gradients are computed using the parameter-shift rule, a gradient evaluation method compatible with quantum circuits. These gradients are integrated into PyTorch's autograd engine, enabling optimization using the standard Adam optimizer. Thus, in HQC-AE, both classical weights and quantum gate parameters are co-trained through unified backpropagation. We use open-source Python and PennyLane [64] for simulating the HQC-AE. The quantum simulator is noise-free, unlike real quantum computers, making it reliable for our analysis.

In each epoch, the validation loss is calculated using (11), and after the training ends, the model with the lowest loss for the validation set is chosen for both AE and HQC-AE. After training, the reconstruction error for each validation sample is computed, and an upper bound threshold is derived using the interquartile range (IQR) method:

$$Threshold = Q_3 + 1.5 \times (Q_3 - Q_1) \qquad (12)$$

where $Q_1$ and $Q_3$ are the first and third quartiles of the validation reconstruction errors. The IQR method in (12) is a statistical technique to identify potential outliers in a dataset. This threshold is then applied to test samples from spoofed scenarios. Any test instance whose reconstruction error exceeds the threshold is flagged as spoofed.

We select the IQR method to ensure that spoofed samples do not remain undetected. Since our detection model is trained exclusively on authentic GNSS signals and has no access to spoofed samples during training, targeting to detect zero-day time-push spoofing attacks, it is not practical to tune the detection threshold based on spoofed data. In other words, it limits the effectiveness of a zero-day attack detection framework if tested on multiple thresholds using the test dataset to select the one that yields the best detection performance. Therefore, we adopt the IQR method under the assumption that spoofed samples will manifest as outliers in terms of reconstruction error. This enables our model to detect spoofed signals without any prior knowledge of the spoofing scenarios, relying solely on training with clean samples. We apply this detection strategy to both the classical AE and the HQC-AE and compare their performance.

VII. ANALYSIS AND RESULTS

*A. Comparison with Classical AE on Zero-Day GNSS Time-Push Spoofing Attack Detection*

Both the classical AE and the HQC-AE are trained using clean static GNSS data and evaluated on the ds2, ds3, and ds7 spoofing scenarios. Each experiment (training and testing) is conducted ten times, and the average results are reported in Table I.

The detection accuracies of AE for ds2, ds3, and ds7 are 93.89%, 94.85%, and 97.78% respectively. In contrast, HQC-AE demonstrates improved performance across all scenarios, achieving accuracies of 97.13%, 97.77%, and 98.23% for ds2, ds3, and ds7, respectively. The average accuracy of AE across all scenarios is 95.51%, while HQC-AE achieves 97.71%, indicating a 2.2% increase in average detection accuracy. These results are illustrated in Fig. 9, which clearly shows that HQC-AE outperforms its classical counterpart for zero-day time-push spoofing attack detection under all scenarios.

TABLE I
PERFORMANCE COMPARISON OF CLASSICAL AE AND HQC-AE ON TEST SAMPLES

| Metrices (%) | ds2 | | ds3 | | ds7 | |
|---|---|---|---|---|---|---|
| | AE | HCQ-AE | AE | HCQ-AE | AE | HCQ-AE |
| Accuracy | 93.89 | 97.13 | 94.85 | 97.77 | 97.78 | 98.23 |
| TPR | 100 | 100 | 100 | 100 | 98.27 | 98.15 |
| FNR | 0 | 0 | 0 | 0 | 1.73 | 1.85 |
| TNR | 69.52 | 85.68 | 76.53 | 89.85 | 95.68 | 98.56 |
| FPR | 30.48 | 14.32 | 23.47 | 10.15 | 4.32 | 1.44 |

At first glance, Fig. 9 may suggest that both AE and HCQ-AE models perform better under the sophisticated spoofing scenario (ds7) compared to the simplistic (ds2) and intermediate (ds3) attacks. However, a closer examination of Table I reveals that the models actually achieve better detection performance on spoofed samples in the simplistic and intermediate scenarios. This is further illustrated in Fig. 10, which presents the true positive rate (TPR) or the spoofed sample detection accuracy. For both AE and HCQ-AE, the TPR is 100% for ds2 and ds3, and above 98% for ds7.

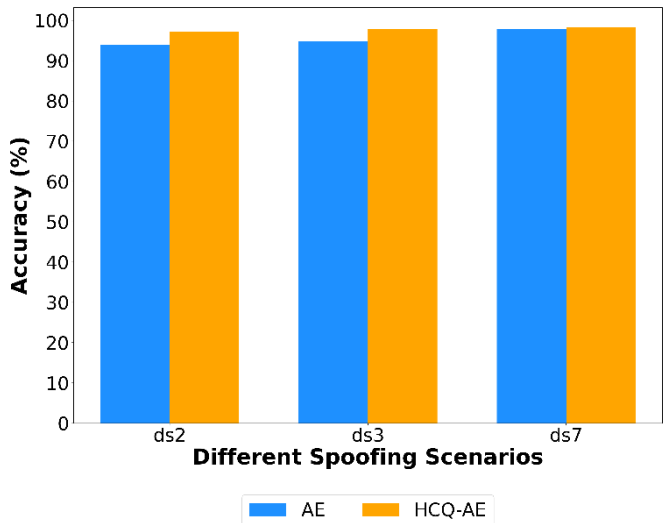

**Fig. 9.** Comparison of zero-day time-push spoofing attack detection accuracy of classical AE and HQC-AE.

The similarity in spoofed sample detection performance between AE and HQC-AE suggests that the overall difference in accuracy arises primarily from their ability to correctly detect clean samples. To highlight this, we analyze the true negative rate (TNR), the clean sample detection accuracy, shown in Fig. 11, based on data from Table I.

As illustrated in Fig. 11, there is a significant performance gap between AE and HQC-AE in detecting clean signals. The TNR of AE for ds2, ds3, and ds7 is 69.52%, 76.53%, and 95.68%, respectively. In comparison, HQC-AE achieves substantially higher TNRs: 85.68% for ds2, 89.85% for ds3, and 98.56% for ds7. This improvement translates to a lower false positive rate (FPR) for HQC-AE, as shown in Fig. 12.

From Table I and Fig. 12, the FPR for AE is 30.48%, 23.47%, and 4.32% for ds2, ds3, and ds7, respectively. For HQC-AE, the FPR is significantly reduced to 14.32%, 10.15%, and 1.44%, respectively. This corresponds to a reduction in false alarm rates by 20.33%, 13.32%, and 2.88% under the ds2, ds3, and ds7 scenarios, respectively. These results clearly demonstrate that HQC-AE outperforms the classical AE, particularly in reducing false alarms while maintaining high spoofed sample detection performance.

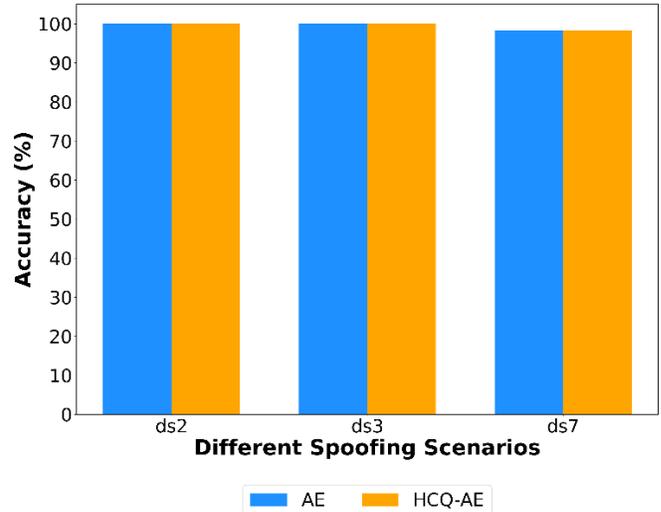

**Fig. 10.** Comparison of spoofed sample detection accuracy (TPR) of classical AE and HQC-AE.

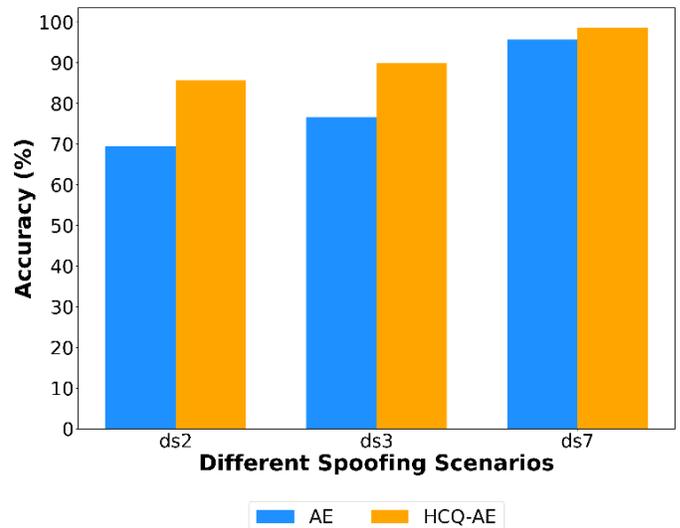

**Fig. 11.** Comparison of clean sample detection accuracy (TNR) of classical AE and HQC-AE.

We also analyze the total number of trainable parameters, including weights and biases, in the neural networks, along with their corresponding storage requirements for both the classical AE and the HQC-AE, as summarized in Table II. The classical AE consists of 12,036 trainable parameters and requires 47.02 kilobytes (KB) of storage. In contrast, the HQC-AE has only 8,132 trainable parameters, with a reduced storage requirement of 31.77 KB. These results demonstrate that the HQC-AE achieves better detection performance while utilizing fewer trainable parameters and requiring less storage than its classical counterpart.

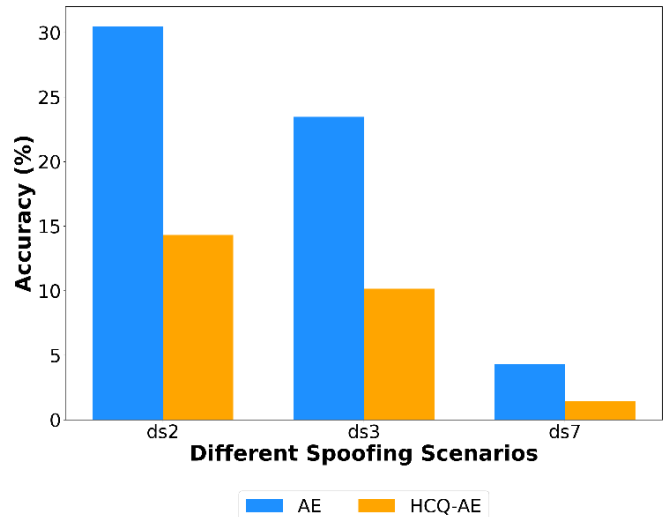

**Fig. 12.** Comparison of false alarm rate (FPR) of classical AE and HQC-AE.

TABLE II
COMPARISON OF THE NUMBER OF MODEL PARAMETERS AND REQUIRED MEMORY SIZE

| Model | Number of Trainable Parameters | Storage Size (KB) |
|---|---|---|
| AE | 12036 | 47.02 |
| **HQC-AE** | **8132** | **31.77** |

*B. Comparison with Traditional Supervised Learning-based ML Models for Unseen GNSS Time-Push Spoofing Attack Detection*

We compare the performance of our HQC-AE with traditional supervised ML models using a leave-one-out strategy. In this approach, each ML model is trained on all scenarios except one and then tested on the excluded scenario. This setup is designed to evaluate how well each model generalizes to unseen spoofing attacks when trained with a combination of clean and known spoofed samples.

As baseline models, we use Random Forest (RF), Support Vector Machine (SVM), K-Nearest Neighbors (KNN), Extreme Gradient Boosting (XGB), and a Neural Network (NN). RF, SVM, and KNN are implemented using the scikit-learn library in Python, while XGB is implemented using the xgboost package in Python. The NN is implemented using PyTorch with an architecture of (input dimension → 64 → 32 → 2), where ReLU is used as the activation function in the hidden layers. The results of this comparison are summarized in Table III.

The results indicate that traditional ML models perform reasonably well on the ds2 and ds3 scenarios but fail to generalize to the more sophisticated spoofing attack in ds7, even when trained on clean static data and spoofed samples from ds2 and ds3. In terms of spoofed sample detection rate (TPR) and the misclassification rate of spoofed samples as authentic clean samples (FNR), our method consistently outperforms traditional ML approaches across all scenarios.

For the subtle ds7 attack, traditional ML models, despite being trained with clean and spoofed data from ds2 and ds3, perform poorly across all evaluation metrics. In contrast, our HQC-AE model, trained solely on clean signals without any prior exposure to spoofed data, achieves 98.23% accuracy, 98.15% TPR, and 1.85% FNR on the ds7 scenario. These results demonstrate that HQC-AE significantly outperforms traditional ML models in detecting previously unseen and sophisticated spoofing attacks, even when those models have been trained on data from other spoofing scenarios, while HQC-AE has only seen clean data during training.

TABLE III
PERFORMANCE COMPARISON WITH TRADITIONAL SUPERVISED ML MODELS FOR UNSEEN GNSS TIME-PUSH SPOOFING ATTACK DETECTION UNDER LEAVE-ONE-OUT CLASSIFICATION STRATEGY

| Metrices | scenario | RF | SVM | KNN | XGB | NN | **HQC-AE** |
|---|---|---|---|---|---|---|---|
| Accuracy (%) | ds2 | **99.95** | **99.95** | **99.95** | **99.95** | **99.95** | 97.13 |
|  | ds3 | **99.86** | 98.84 | 89.38 | **99.86** | 94.99 | 97.77 |
|  | ds7 | 56.55 | 36.78 | 37.64 | 29.61 | 33.50 | **98.23** |
| TPR (%) | ds2 | 100 | 100 | 100 | 100 | 100 | **100** |
|  | ds3 | 99.82 | 98.51 | 86.93 | 99.82 | 94.36 | **100** |
|  | ds7 | 46.42 | 22.04 | 23.10 | 13.20 | 17.95 | **98.15** |
| FNR (%) | ds2 | 0 | 0 | 0 | 0 | 0 | **0** |
|  | ds3 | 0.18 | 1.49 | 13.07 | 0.18 | 5.64 | **0** |
|  | ds7 | 53.58 | 77.96 | 76.90 | 86.80 | 82.05 | **1.85** |

*C. Comparison with Existing Work*

We compare the performance of our HQC-AE with recent work. First, we compare our results with the method developed by Iqbal et al. in [53], which introduces an unsupervised, representation learning-based spoofing detection approach. We selected this work for comparison because it, like ours, focuses on detecting unseen GNSS spoofing attacks using a model trained exclusively on clean GNSS signals, in a zero-day detection fashion. Additionally, Iqbal et al. also used the TEXBAT dataset for their evaluation, aligning with our experimental setup.

The authors employ a variational autoencoder (VAE) trained with clean, unspoofed samples and evaluated across various spoofing scenarios [53]. Their method uses different thresholds for reconstruction loss based on acceptable FPR, where an acceptable threshold is often application-specific. They conduct a sensitivity analysis for different acceptable FPR, from which we choose the results providing balanced performance across all attacked scenarios. We compare their results under the same test scenarios, ds2, ds3, and ds7, with our HQC-AE, as summarized in Table IV. Here, for [53], we present the results reported by the authors of the study, which were obtained using a different set of features identified as the best for their method. Therefore, we directly present their results rather than rerunning the experiments with our selected features.

The results show that our method provides similar performances under ds2 and ds3 in terms of overall accuracy and TPR. Although our method exhibits higher FPR in these scenarios, it achieves perfect spoofed sample detection (TPR = 100%), which is critical for spoofing resilience. For ds7, the most challenging spoofing attack in TEXBAT, our method significantly outperforms the VAE-based approach across all key metrics, including overall accuracy, TPR, and FPR.

TABLE IV
PERFORMANCE COMPARISON WITH RECENT WORK

| Metrics (%) | Attack Scenarios | | | | | |
|---|---|---|---|---|---|---|
| | ds2 | | ds3 | | ds7 | |
| | Iqbal Et al. [53] | HQC-AE | Iqbal Et al. [53] | HQC-AE | Iqbal Et al. [53] | HQC-AE |
| Accuracy | **97.90** | 97.13 | 97.72 | **97.77** | 92.49 | **98.23** |
| TPR | 99.99 | **100** | 100 | **100** | 92.50 | **98.15** |
| FNR | 0.01 | **0** | 0 | **0** | 7.50 | **1.85** |

Additionally, Iqbal et al. [53], compared their approach against several recent unsupervised spoofing detection techniques under the ds7 scenario, which is widely regarded as the most challenging in the TEXBAT dataset due to its high stealth and minimal observable anomalies. To further validate our method, we compare the performance of our HQC-AE with those same techniques, using the results reported in [53]. The comparison, summarized in Table V, demonstrates that HQC-AE achieves superior performance across all evaluation metrics. Similarly, for the comparison in Table V, we present the results, which were obtained using different sets of features identified as the best for their respective methods in [16], [43], [47], [61]. Therefore, we directly present their results rather than rerunning the experiments with our selected features.

TABLE V
PERFORMANCE COMPARISON WITH EXISTING WORK ON UNSUPERVISED LEARNING FOR THE TEXBAT DS7 SCENARIO

| Methods | Accuracy | TPR | FNR | FPR |
|---|---|---|---|---|
| Wesson et al. [16] | 71.25 | 73.96 | 26.04 | 8.56 |
| Manfredini et al. [47] | 69.35 | 68.34 | 31.66 | 11.68 |
| Sun et al. [43] | 56.34 | 57.68 | 42.32 | 8.64 |
| Zhou et al. [61] | 78.55 | 81.65 | 18.35 | 9.32 |
| **HQC-AE** | **98.23** | **98.15** | **1.85** | **1.44** |

Under the sophisticated spoofing scenario ds7, HQC-AE achieves 98.23% accuracy, which represents an improvement of approximately 38% over Wesson et al. [16], 42% over Manfredini et al. [47], 74% over Sun et al. [43], and 25% over Zhou et al. [61]. This gain means HQC-AE makes correct decisions far more often, substantially reducing classification errors. For TPR, HQC-AE reaches 98.15%, improving over the same baselines by about 33%, 44%, 70%, and 20%, respectively, which indicates a much higher proportion of spoofing attempts are detected before causing harm. The FNR drops to 1.85%, which is a reduction of roughly 93%, 94%, 96%, and 90% from the prior methods, meaning the likelihood of missing an attack is almost eliminated. Similarly, the FPR is 1.44%, reduced by around 83%, 88%, 83%, and 85%, respectively, translating into far fewer false alarms and minimal disruption to normal operations. These percentage improvements show that HQC-AE not only increases the rate of correctly identifying both spoofed and clean signals but also greatly reduces the rates of missed detections and false alarms. In practical spoofing defense, the large FNR reduction is the most critical, as preventing undetected attacks is more important than eliminating every false alarm, but the simultaneous improvement across all four metrics demonstrates HQC-AE's balanced and robust performance in high-threat environments.

## VIII. CONCLUSIONS

GNSS plays a critical role in ensuring precise time synchronization in different applications. Time-push spoofing attacks targeting GNSS static receivers can lead to catastrophic failures, potentially causing widespread disruptions, particularly in IoT environments, where interconnected systems amplify the impact of a single point of failure. To address this threat, we present an HQC-AE-based zero-day time-push GNSS spoofing attack detection approach. Our model is trained exclusively on clean GNSS signals and does not require prior exposure to spoofed data. By integrating quantum layers with classical neural networks, the HQC-AE leverages the enhanced representational capabilities of hybrid quantum-classical learning, potentially offering a quantum advantage in feature extraction from clean signals. We evaluate our approach using the TEXBAT benchmark across all levels of time-push spoofing complexity, simplistic, intermediate, and sophisticated, for static receivers. HQC-AE consistently outperforms its classical counterpart, the classical autoencoder (AE), in zero-day detection of time-push spoofing attacks.

Additionally, we compare HQC-AE with traditional supervised ML models using a leave-one-out classification strategy, where the ML models are trained on both clean and spoofed data while withholding one spoofing scenario for testing. Our results show that HQC-AE outperforms these models, particularly in detecting unseen and sophisticated attacks, despite being trained solely on clean data. Furthermore, we benchmark our model against recent state-of-the-art approaches for detecting subtle attacks, the ds7 scenario in TEXBAT. Our HQC-AE achieves superior performance, demonstrating its robustness and generalization capability. In future work, we aim to extend our detection framework to include both static and dynamic receiver configurations, covering a wider spectrum of spoofing attack types, including both time-push and position-push attacks, using advanced hybrid quantum-classical learning techniques to further enhance spoofing attack detection accuracy and reliability.

## ACKNOWLEDGMENT

This work is based upon the work supported by the National Center for Transportation Cybersecurity and Resiliency (TraCR) (a US Department of Transportation National University Transportation Center) headquartered at Clemson

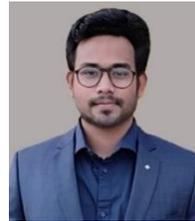

**Abyad Enan** (Member, IEEE) is a Ph.D. student at Clemson University, Clemson, SC, USA. He received his M.S. in Civil Engineering from Clemson University in 2024 and a B.S. in Electrical and Electronic Engineering from Bangladesh University of Engineering and Technology (BUET), Dhaka, Bangladesh, in 2018. His research interests include transportation cyber-physical systems, cybersecurity, and quantum computing.

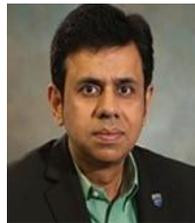

**Mashrur Chowdhury** (Senior Member, IEEE) received his Ph.D. degree in civil engineering from the University of Virginia in 1995. He is the Eugene Douglas Mays Chair of Transportation in the Glenn Department of Civil Engineering, Clemson University, SC, USA. He is the Founding Director of the USDOT-sponsored USDOT National Center for Transportation Cybersecurity and Resiliency (TraCR). He is also the Director of the Complex Systems, Data Analytics and Visualization Institute (CSAVI) at Clemson University. Dr. Chowdhury is a Registered Professional Engineer in Ohio, USA. He is a Fellow of the American Society of Civil Engineers and a Senior Member of IEEE.

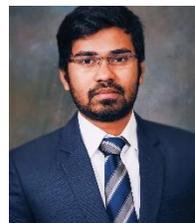

**Sagar Dasgupta** received his B.Tech. degree in Mechanical Engineering from Motilal Nehru National Institute of Technology Allahabad, Prayagraj, India, in 2015, M.S. degree in Mechanical Engineering from Clemson University, Clemson, SC, USA, in 2020, and Ph.D. degree in Civil Engineering from The University of Alabama, Tuscaloosa, AL, USA, in 2024. His research interests include GNSS-based positioning and navigation security, transportation cyber-physical system security, transportation digital twin, and intelligent transportation systems.

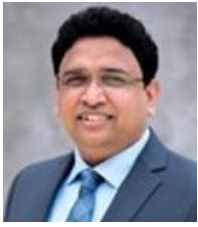 **Mizanur Rahman** (Senior Member, IEEE) received his M.Sc. and Ph.D. degrees in Civil Engineering (transportation systems) from Clemson University, Clemson, SC, USA, in 2013 and 2018, respectively. He is an Assistant Professor in the Department of Civil, Construction, and Environmental Engineering at the University of Alabama (UA), Tuscaloosa, AL, USA. His research focuses on cybersecurity and digital twins.